# Worm-level Control through Search-based Reinforcement Learning

**Mathias Lechner**[*]
TU Wien
Austria

**Radu Grosu**
TU Wien
Austria

**Ramin M. Hasani**
TU Wien
Austria

## Abstract

Through natural evolution, nervous systems of organisms formed near-optimal structures to express behavior. Here, we propose an effective way to create control agents, by *re-purposing* the function of biological neural circuit models, to govern similar real world applications. We model the tap-withdrawal (TW) neural circuit of the nematode, *C. elegans*, a circuit responsible for the worm's reflexive response to external mechanical touch stimulations, and learn its synaptic and neural parameters as a policy for controlling the inverted pendulum problem. For reconfiguration of the purpose of the TW neural circuit, we manipulate a search-based reinforcement learning. We show that our neural policy performs as good as existing traditional control theory and machine learning approaches. A video demonstration of the performance of our method can be accessed at `https://youtu.be/o-Ia5IVyff8`.

## Introduction

The nervous system of the soil-worm, C. elegans, has entirely been mapped, demonstrating a near-optimal wiring structure [13]. An adult hermaphrodite's brain is composed of 302 neurons hard-wired by around 8000 chemical and electrical synapses [2]. Function of many neural circuits within its brain has been identified [14, 1, 4, 6]. In particular, a neural circuit which is responsible for inducing a forward/backward locomotion reflex when the worm is mechanically exposed to touch stimulus on its body, has been well-characterized [1] (see Fig. 1A for an illustration of this reflex). Synaptic polarities of the circuit have then been predicted, suggesting that the circuit realizes a competitive behavior between forward and backward reflexes, in presence of touch stimulations [14, 15].

Behavior of the tap-withdrawal (TW) reflexive response is substantially similar to the impulse response of a controller operating on an *Inverted Pendulum* [9] dynamic system, as illustrated in Fig. 1A and 1B. We thought of taking advantage of such similarity and reconfigure the synaptic and neural parameters of a deterministic model of the TW neural circuit, to control the inverted pendulum within the OpenAI's Roboschool environment [11], while preserving the near-optimal wiring structure of the circuit. We deploy a search-based reinforcement learning (RL) algorithm for synaptic parametrization and discuss the performance of the TW re-purposed network, throughly.

## Methods

In this section we first define how the TW circuit interacts with the inverted pendulum environment and then throughly discuss the parameter optimization of the circuit by a search-based RL algorithm.

---

[*]Correspondence to mathias.lechner@hotmail.com

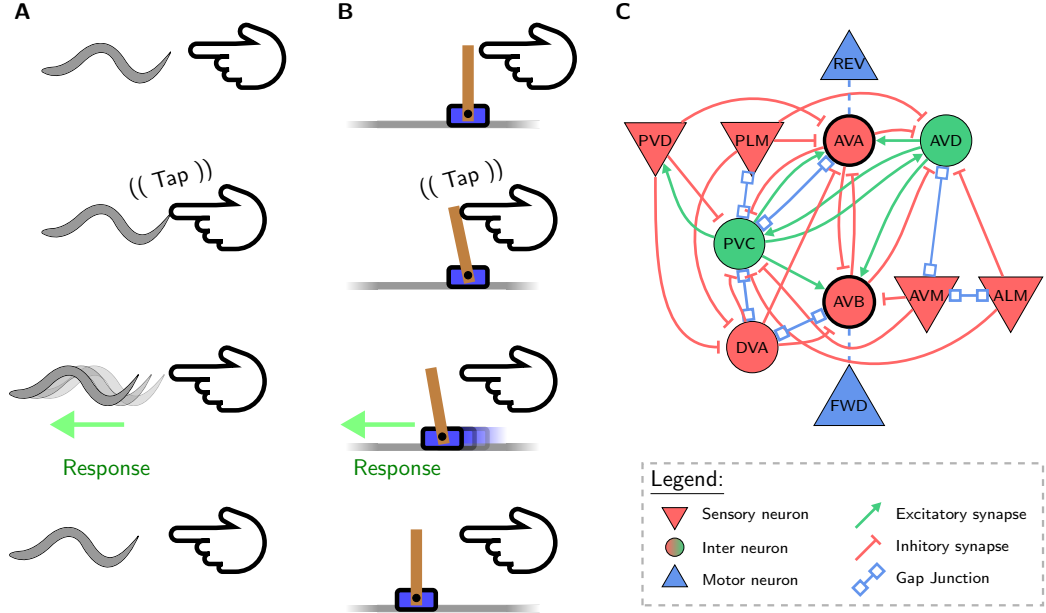

Figure 1: Illustration of the touch withdrawal reflex. **A)** Touching the worm's tail will excite the touch sensory neuron PLM, and correspondingly induces a forward locomotion command in the animal [1] (Hand is not drawn to scale!). **B)** Working principle of the introduced touch withdrawal inverted-pendulum controller. **C)** Tap Withdrawal neural circuit of *C. elegans*.

**Manipulating the model of the TW neural circuit for the inverted pendulum swingup problem**

We first introduce the neuron and synapse models we utilized to make an artificial TW neural circuit. Neurons are modeled by the well-known, deterministic ordinary differential equation (ODE), *Leaky-integrate-and-fire* (LIF) model [3] as $C_m \frac{dv_i}{dt} = G_{Leak}\left(V_{Leak} - v_i(t)\right) + \sum_{i=1}^{n} I_{in}^{(i)}$, where $C_m, G_{Leak}$ and $V_{Leak}$ are parameters of the neuron and $I_{in}^{(i)}$, stands for the external currents to the cell.

Chemical synapses are where two neurons trade information by the release of neurotransmitters. Chemical synaptic currents are calculated by a non-linear component standing for their conductance strength, which exponentially depends on the presynaptic neurons' potential, $V_{pre}$, as $g(V_{pre}) = w/1 + e^{\sigma(V_{pre}+\mu)}$. The synapse model adopted from [5]. These currents linearly depend on the postsynaptic neuron's membrane potential as well, as $I_s = g(V_{pre})(E - V_{post})$, where by varying E, they can make inhibitory or excitatory connection to their postsynaptic neurons. An electrical synapse (gap-junction), which is a physical junction between two neurons, is modeled by a constant conductance model, where based on the Ohm's law their bidirectional current between neurons $j$ and $i$, can be computed as $\hat{I}_{i,j} = \hat{\omega}\left(v_j(t) - v_i(t)\right)$.

To simulate a neural circuit we used a fixed step *implicit* numerical ODE solver [7].

For interacting with environment, We introduce sensory and motor neurons. A sensory component consists of two neurons $S_p, S_n$ and a measurable dynamic system variable $x$. $S_p$ gets activated when $x$ has a positive value, whereas $S_n$ fires when $x$ is negative. Mathematically, the potential of the

| Environment variable | Type | Positive neuron | Negative neuron |
|---|---|---|---|
| $\varphi$ | Observation | PLM | AVM |
| $\dot{\varphi}$ | Observation | ALM | PVD |
| $a$ | Action | FWD | REV |

Table 1: Mapping of environment variables to sensory and motor neurons

neurons $S_p$, and $S_n$, as a function of $x$, can be expressed as

$$S_p(x) := \begin{cases} -70mV & \text{if } x \leq 0 \\ -70mV + \frac{50mV}{x_{max}}x & \text{if } 0 < x \leq x_{max} \\ -20mV & \text{if } x > x_{max} \end{cases} \quad (1)$$

$$S_n(x) := \begin{cases} -70mV & \text{if } x \geq 0 \\ -70mV + \frac{50mV}{x_{min}}x & \text{if } 0 > x \geq x_{min} \\ -20mV & \text{if } x < x_{min}. \end{cases} \quad (2)$$

This maps the region $[x_{min}, x_{max}]$ of system variable $x$, to a membrane potential range of $[-70mV, -20mV]$. Note that the potential range is selected to be close to the biophysics of the nerve cells where the resting potential is usually set around -70 mV and a neuron can be considered to be active when it has a potential around -20 mV [5]. The TW neural circuit shown in Figure (1C) has four sensory neurons: PVD, PLM, AVA and ALM. It therefore, allows us to map the circuit to two dynamic system variables. The inverted pendulum environment provides four observation variables: The position of the cart $x$, together with its velocity $\dot{x}$, the angle of the pendulum $\varphi$[1] along with its angular velocity $\dot{\varphi}$. Thus in order to map our TW circuit model to the environment, a compromise that selects two out of these four observation variables has to be found.

Since the objective of the inverted pendulum task is to control the angle in an upward way, we decided to feed $\varphi$ and $\dot{\varphi}$ as the inputs to the TW circuit. Consequently, the controller circuit is not aware of where the cart is or whether the boundary of the movable space is reached. two input variables are mapped to the circuit elements as shown in Table 1.

Similar to sensory neurons, a motor component is comprised of two neurons $M_n$, $M_p$ and a controllable motor variable $y$. Values of $y$ is computed by $y := y_p + y_n$ and

$$y_p(M_p) := \begin{cases} y_{max} & \text{if } M_p > -20mV \\ \frac{y_{max}}{50mV}(M_p + 70mV) & \text{if } -70mV < M_p \leq -20mV \\ 0 & \text{if } M_p < -70mV \end{cases} \quad (3)$$

$$y_n(M_n) := \begin{cases} y_{min} & \text{if } M_n > -20mV \\ \frac{y_{min}}{50mV}(M_n + 70mV) & \text{if } -70mV < Mn \leq -20mV \\ 0 & \text{if } M_n < -70mV \end{cases} \quad (4)$$

This maps the neuron potentials $M_n$ and $M_p$, to the range $[y_{min}, y_{max}]$. The TW circuit in Fig. (1C) contains two abstract representation for motor neurons involved in reversal movement as REV, and the ones functioning in forward locomotion, FWD. These neurons are allocated to the motor command of the cart in the inverted pendulum environment, as expressed in Table 1.

**Search-based reinforcement learning**

In this section we formulate a RL setting for training the parameters of the neural circuit to perform the swingup of the inverted pendulum.

The behavior of a neural circuit can be expressed as a policy $\pi_\theta(o_i, s_i) \mapsto \langle a_{i+1}, s_{i+1} \rangle$, that maps an observation $o_i$, and an internal state $s_i$, to an action $a_{i+1}$, and a new internal state $s_{i+1}$. This policy acts upon a possible stochastic environment $Env(a_{i+1})$, that provides an observation $o_{i+1}$, and reward $r_{i+1}$, feedback. The stochastic return is given by $R(\theta) := \sum_{t=1}^{T} r_t$. Objective of the *Reinforcement learning* is to find a $\theta$ that maximizes $\mathbb{E}(R(\theta))$.

---
[1]Remark: The environment further splits $\varphi$ into $sin(\varphi)$ and $cos(\varphi)$ to avoid the $2\pi \to 0$ discontinuity

| Parameter | Min value | Max value |
|---|---|---|
| $C_m$ | 1mF | 1F |
| $G_{Leak}$ | 50mS | 5S |
| $E_{rev}$ excitatory | 0mV | 0mV |
| $E_{rev}$ inhibitory | -90mV | -90mV |
| $V_{Leak}$ | -90mV | 0mV |
| $\mu$ | -40mV | -40mV |
| $\sigma$ | 0.05 | 0.5 |
| $\omega$ | 0S | 3S |
| $\hat{\omega}$ | 0S | 3S |

Table 2: Table of parameters and their boundaries of a neural circuit

**Algorithm 1:** Random Search with Decaying Objective Indicator

**Input:** A stochastic objective indicator $f$, a starting parameter $\theta$
**Output:** Optimized parameter $\theta$

1  $f_\theta \leftarrow f(\theta)$;
2  **for** $k \leftarrow 1$ **to** *maximum iterations* **do**
3    $\quad \theta' \leftarrow \theta + rand()$;
4    $\quad f_{\theta'} \leftarrow f(\theta')$;
5    $\quad$ **if** $f_{\theta'} < f_\theta$ **then**
6    $\quad\quad$ Set $\theta \leftarrow \theta'$;
7    $\quad\quad f_\theta \leftarrow f_{\theta'}$;
8    $\quad\quad i \leftarrow 0$;
9    $\quad$ **end**
10   $\quad i \leftarrow i + 1$;
11   $\quad$ **if** $i > N$ **then**
12   $\quad\quad f_\theta \leftarrow f(\theta)$;
13   $\quad$ **end**
14 **end**
15 **return** $\theta$;

Approaches to find such optimal $\theta$, can be categorized into two major sections, based on how randomness is formulated for the environment explorations [10]: I-Gradient-based and II-search-based methods.

Gradient-based RL performs initial random sampling for generating $a_i$, and uses action's influence on the return value, to update $\theta$ [16]. Search-based methods randomly sample parameters and evaluate how good they perform to update $\theta$ [10]. Here, we adopt a search-based optimization as these methods can be applied regardless of the internal structure of the policy (*black-box optimization*).

One major obstacle for search-based optimization is stochastic nature of the objective function $R(\theta)$, which makes this problem an instance of *Stochastic Optimization* [12]. A possible solution to overcome the high variance of $R(\theta)$, is to rely on a very large number of samples [10], which nonetheless comes with high computational costs.

Our approach is based upon a *Random Search* (RS) [8] optimization, combined with an *Objective Estimation* (OE) as objective function $f : \theta \mapsto \mathbb{R}^+$. The OE exploits the fact that a good policy $\pi_\theta$, should perform well even in challenging conditions (worst cases). The return of one $\theta$, is evaluated $N$ times, and the average of the $k$ lowest returns is used as OE. Empirical evaluations confirmed that such estimation scheme performs better than taking the average or the lowest value of $N$ samples as objective function, to optimize the TW neural circuit. To ensure that a single spuriously high OE for some $\theta$ does not hinder the algorithm to find a legitimately good parameter, the OE must be reevaluated after it is utilized $M$ times, for comparisons with other OEs.

## Results and Discussions

We set up the search-based RL algorithm to optimize the parameters $\omega, \hat{\omega}, \sigma, C_m, V_{Leak}$ and $G_{Leak}$ of each neuron and synapse of the TW neural circuit shown in Fig. (1) in the RoboSchool[11] `RoboschoolInvertedPendulum-v1` environment. The termination condition of the procedure was set to a fixed runtime of 12 hours. The final TW neural circuit achieves the maximum possible return of 1,000, which is equal to the performance of a PID-state-controller or an artificial neural net [11]. However, as the position and velocity of the cart are either feed into the circuit or affect the reward directly, the controlled cart is subject to a small drift, that will make it reach one end of the movable space after the evaluate of the environment has ended. A video demonstration of the TW controller can be found at `https://youtu.be/o-Ia5IVyff8`.

In this work We showed that a model of a neural circuit from a biological organism can be adapted to function as a controller for a task with similar characteristics to the original purpose of the circuit. We will explore capabilities of such circuits in a future work.

**Acknowledgments**

This work was supported with computation resources by Microsoft Azure via the Microsoft Azure for Research Award program.